%% file: main.tex
\pdfoutput=1

\documentclass[letterpaper, preprint, paper,11pt]{AAS}	

\usepackage{bm}
\usepackage{amsmath}
\usepackage{subfigure}
\usepackage[colorlinks=true, pdfstartview=FitV, linkcolor=black, citecolor= black, urlcolor= black]{hyperref}
\usepackage{overcite}
\usepackage{footnpag}			      	
\usepackage{amssymb}

\PaperNumber{26-740}

\begin{document}

\title{PPO vs. SAC: A Comparative Study for Cooperative Motion Control of Dual-Arm Space Robots}

\author{Nasim Fallahiarezoodar\thanks{PhD Candidate, Department of Earth and Space Science and Engineering, York University, M3J1P3, Student Member of AIAA},
Zheng H. Zhu\thanks{Professor \& Tier 1 Research Chair, Mechanical Engineering, York University, M3J1P3, AIAA Associate Fellow},
}

\maketitle{}

\input{00_abstract}
\input{01_Intro}
\input{02_ProposedSystem}
\input{03_TechnicalPreliminaries}
\input{04_ProblemFormulation}
\input{05_SimulationResults}
\input{06_Conclusion}
\input{07_Acknowledgment}

\bibliographystyle{AAS_publication}   
\bibliography{references}   

\end{document}

%% file: 00_abstract.tex
\begin{abstract}
The growing population of orbital debris has intensified interest in active debris removal and on-orbit servicing, for which dual-arm space manipulators offer improved grasping stability and load distribution relative to single-arm systems. However, coordinating two manipulators under inter-arm collision avoidance, orientation alignment, and sensing uncertainty remains challenging. This paper develops a coordinated reinforcement learning framework for cooperative motion planning of a dual-arm space manipulator in a target-reaching task with inter-arm collision-avoidance constraints. A centralized state representation combines joint-space, end-effector, and target-relative information, while a multi-objective reward function promotes target-reaching accuracy, orientation alignment, motion smoothness, synchronized task completion, and collision avoidance. Within this framework, Proximal Policy Optimization (PPO) and Soft Actor-Critic (SAC) are compared as representative on-policy and off-policy algorithms, and the learned policies are further evaluated for generalization to previously unseen target configurations. Across five random seeds, SAC substantially outperforms PPO within the considered fixed-base simulation scenarios: it reaches an 80\% success threshold in every seed (median 1.6M environment steps) whereas PPO does not reach it within a 4M-step budget, and it attains a higher mean success rate (78.6\% versus 7.4\%) with lower position and orientation errors and tighter bimanual synchronization. Both algorithms degrade on unseen boundary targets, but SAC retains most of its performance while PPO's collapses. The results also expose a failure mode of PPO: under the two-phase shaped reward, it tends to converge to a reward-loitering local optimum rather than committing to the precision task. This behavior highlights the sensitivity of on-policy learning to reward shaping.
\end{abstract}

%% file: 01_Intro.tex
\section{Introduction}
\label{sec:introduction}

\subsection{Background and Motivation}
Orbital debris now poses a major threat to the long-term sustainability, safety, and economic viability of space operations. The accumulation of non-functional satellites, spent launch-vehicle stages, and fragmentation debris has substantially increased the collision risk in low Earth orbit, where a single high-energy collision could trigger a cascading fragmentation process that compromises the usability of critical orbital regions. These concerns have brought active debris removal, on-orbit servicing, and autonomous capture to the forefront of space-safety research.~\cite{flores2014review,fallahiarezoodar2025review}

Robotic manipulators have emerged as one of the most practical enabling technologies for these tasks. Unlike human-crewed approaches, they interact directly with non-cooperative targets, perform capture and stabilization maneuvers, and operate without exposing astronauts to risk. Dual-arm configurations are particularly promising, since coordinated manipulation distributes contact forces more effectively, improves the ability to secure irregular or tumbling objects, and reduces the likelihood of grasp failure relative to single-arm systems.~\cite{alizadeh2024comprehensive} These advantages are especially important when the target provides no dedicated grapple fixture and its motion is only partially known.

However, coordinating two manipulators in orbit is difficult. Inter-arm collision avoidance, kinematic redundancy, and the need for smooth, synchronized motion under sensing uncertainty add substantial complexity. Meeting these demands reliably is therefore central to making dual-arm space manipulators practically viable.

\subsection{Related Work}
Classical nonlinear, robust, and optimal control strategies have been applied to many dynamical systems to mitigate disturbances and modeling uncertainties.~\cite{shi2022robust,11129913,10801815,wang2022robust,WANG202611355,XIE2020156,fallahiarezoodar2024optimal} However, these methods typically rely on accurate models and carefully tuned controllers, which limits their adaptability to the highly uncertain, contact-rich capture scenarios encountered in dual-arm operations involving non-cooperative targets.

These limitations motivate learning-based approaches, and in particular deep reinforcement learning (DRL), for cooperative motion control of dual-arm space robots. DRL methods can learn coordination policies directly from interaction data, capturing complex contact dynamics, uncertainty, and redundancy without requiring explicit analytic controller design. Among DRL methods, Proximal Policy Optimization (PPO) and Soft Actor-Critic (SAC) represent two widely used paradigms with distinct learning mechanisms, sample-efficiency properties, and exploration strategies.

PPO has attracted considerable attention in space-robotic motion planning owing to its stable policy updates, implementation simplicity, and suitability for continuous control. Wang et al. applied a PPO-based strategy to multi-target trajectory planning for a six-degree-of-freedom (6-DoF) free-floating space robot, demonstrating improved tracking performance and robustness under external disturbances, base-parameter variations, and joint-failure scenarios.~\cite{wang2021multi} Reinforcement learning has also been investigated for dual-arm robotic capture of non-cooperative dynamic targets, where sim-to-real transfer improved the practical applicability of the learned capture policies.~\cite{du2023reinforcement} More recent studies have extended learning-based motion planning toward dual-arm free-floating space robots and non-cooperative target capture, where coordinated planning must address dynamic coupling, high-dimensional action spaces, and end-effector pose constraints.~\cite{cao2023reinforcement,wang2022learning} Nevertheless, because PPO is an on-policy algorithm, it often requires substantial interaction data, and its performance can be sensitive to reward design, network architecture, and hyperparameter tuning.

In contrast, off-policy actor--critic methods offer a promising alternative for improving sample efficiency in continuous control. SAC, in particular, incorporates entropy regularization to promote exploration and stabilize learning, making it well suited to high-dimensional continuous action spaces. SAC-based approaches have been applied to robotic path planning and obstacle avoidance, including dynamic obstacle avoidance, multi-arm manipulator path planning, and dual-arm trajectory planning in configuration space.~\cite{prianto2020path,cao2023reinforcement,choi2021reinforcement,li2025deep}

The most closely related space-robot study jointly incorporates collision avoidance, orientation alignment, and observation-noise robustness for a free-floating manipulator, but does so for a \emph{single} arm using Deep Deterministic Policy Gradient (DDPG).~\cite{alali2024rlpath} More broadly, prior dual-arm space-robot studies each commit to a single learning algorithm, predominantly off-policy DDPG~\cite{wu2020reinforcement,li2021constrained} or PPO alone~\cite{karim2025davil}, and none contrast on-policy PPO with maximum-entropy SAC under an identical dual-arm formulation.

\subsection{Contributions}
Despite this progress, existing learning-based studies rarely address, within a single dual-arm space-manipulator formulation, the joint requirements of target-reaching accuracy, inter-arm collision avoidance, orientation alignment, motion smoothness, and synchronized task completion, while also generalizing to unseen target configurations. Direct comparisons between on-policy and off-policy algorithms under an identical dual-arm task formulation also remain limited, as summarized in Table~\ref{tab:related_work_matrix}. To address this limitation, this paper develops a coordinated reinforcement learning framework for cooperative motion planning of a dual-arm space manipulator and evaluates PPO and SAC within the same formulation.

\begin{table}[htbp]
	\fontsize{9}{11}\selectfont
	\centering
	\caption{Objective coverage of representative deep reinforcement learning studies for space and dual-arm manipulators. \checkmark~= addressed; --~= not addressed. Reach: target reaching; Coll: collision avoidance; Orient: orientation alignment; Smooth: motion smoothness; Sync: synchronized bimanual completion; Noise: observation-noise robustness; Gen: generalization to unseen targets. To our knowledge, no prior work jointly addresses this objective set within a single dual-arm space-manipulator formulation.}
	\label{tab:related_work_matrix}
	\begin{tabular}{l c c c c c c c c}
		\hline
		Study & Reach & Coll & Orient & Smooth & Sync & Noise & Gen & Algorithm \\
		\hline
		Wu et al.~\cite{wu2020reinforcement}   & \checkmark & --         & --         & --         & --         & --         & --         & DDPG \\
		Li et al.~\cite{li2021constrained}     & \checkmark & \checkmark & --         & \checkmark & --         & --         & --         & DDPG \\
		DA-VIL~\cite{karim2025davil}           & \checkmark & --         & --         & --         & \checkmark & --         & --         & PPO \\
		Al Ali \& Zhu~\cite{alali2024rlpath}   & \checkmark & \checkmark & \checkmark & --         & --         & \checkmark & --         & DDPG \\
		This work                              & \checkmark & \checkmark & \checkmark & \checkmark & \checkmark & --         & \checkmark & PPO, SAC \\
		\hline
	\end{tabular}
\end{table}

Rather than focusing solely on algorithm selection, the proposed approach constructs a task-specific cooperative learning architecture tailored to space-robotic capture operations. The objective is to enable both manipulators to reach their assigned targets while maintaining safe inter-arm motion, satisfying task-level position and orientation requirements, and producing smooth, physically feasible joint-space trajectories. To this end, a centralized state representation combines joint-space states, end-effector kinematics, and target-relative observations, and a multi-objective reward function balances target-reaching accuracy, orientation alignment, motion smoothness, inter-arm collision avoidance, and synchronized task completion. Within this framework, PPO and SAC are employed as representative on-policy and off-policy algorithms to evaluate how different policy-optimization paradigms realize the proposed formulation.

The main contributions are summarized as follows:
\begin{itemize}
    \item A reinforcement-learning-based formulation is developed for cooperative motion planning of a dual-arm space manipulator in a target-reaching task with inter-arm collision-avoidance constraints.

    \item A centralized state representation is constructed by incorporating joint-space information, end-effector kinematics, and target-relative observations to support coordinated dual-arm policy learning.

    \item A multi-objective reward function is designed to promote target-reaching accuracy, orientation alignment, smooth joint motion, synchronized task completion, and inter-arm collision avoidance.

    \item PPO and SAC are compared as representative on-policy and off-policy DRL algorithms to assess their convergence behavior, task performance, and suitability for the proposed cooperative motion-planning problem.

    \item The learned policies are evaluated on unseen target configurations to examine their generalization capability within the considered simulation scenarios.
\end{itemize}

The remainder of this paper is organized as follows. Section~\ref{sec:proposed_system} describes the dual-arm robotic system and its reachable workspace. Section~\ref{sec:technical_preliminaries} reviews the reinforcement learning preliminaries and the PPO and SAC algorithms. Section~\ref{sec:problem_formulation} formulates the cooperative motion-planning problem as a Markov decision process, including the state and action spaces and the multi-objective reward design. Section~\ref{sec:simulation_results} presents the comparative simulation results for PPO and SAC. Finally, Section~\ref{sec:conclusion} concludes the paper and outlines future work.

%% file: 02_ProposedSystem.tex
\section{Proposed System}
\label{sec:proposed_system}
\begin{figure}[htb]
	\centering
    \begin{minipage}[t]{0.47\linewidth}
        \centering
        \includegraphics[
            width=\linewidth,
            trim={0.0cm 5.0cm 0cm 0.0cm},
            clip
        ]{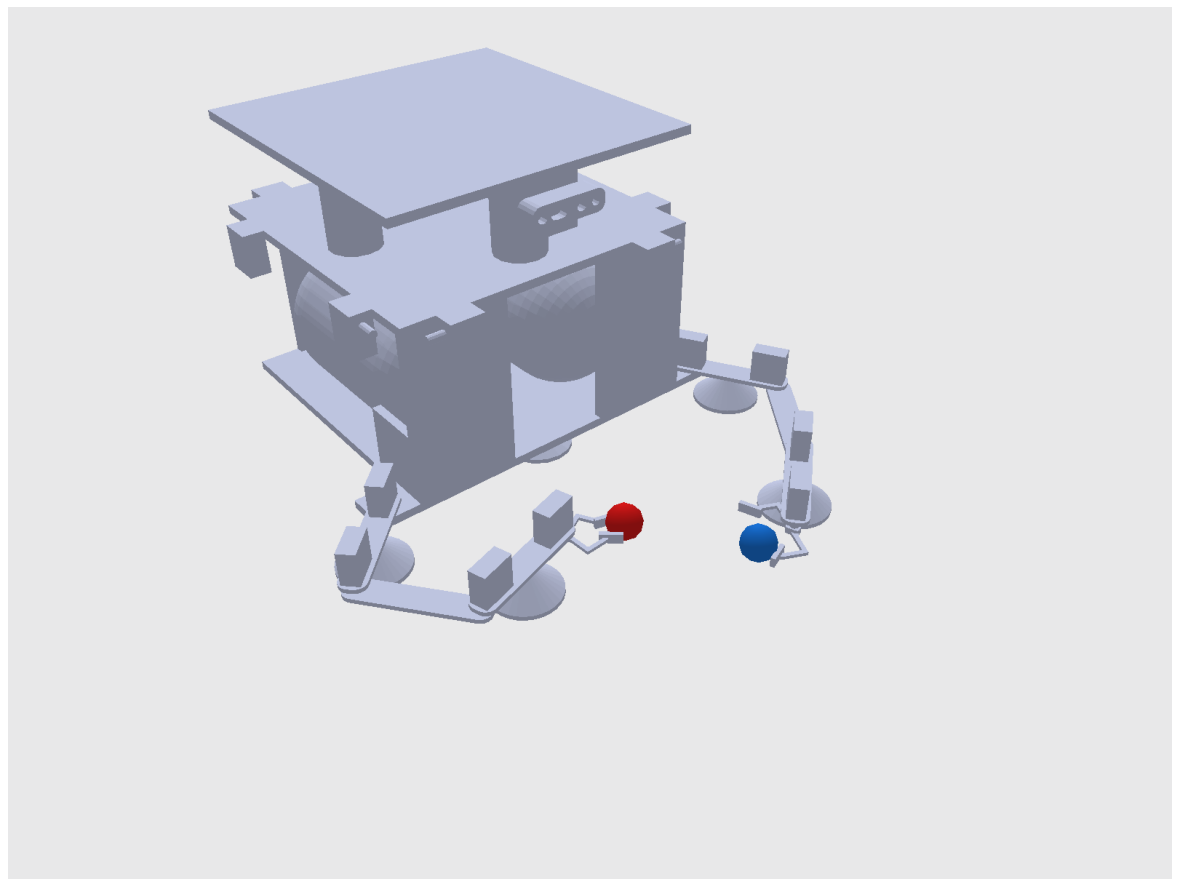}\\[2pt]
        (a)
    \end{minipage}
    \hfill
    \begin{minipage}[t]{0.47\linewidth}
        \centering
        \includegraphics[
            width=\linewidth,
            trim={0.0cm 5.0cm 0.0cm 0.0cm},
            clip
        ]{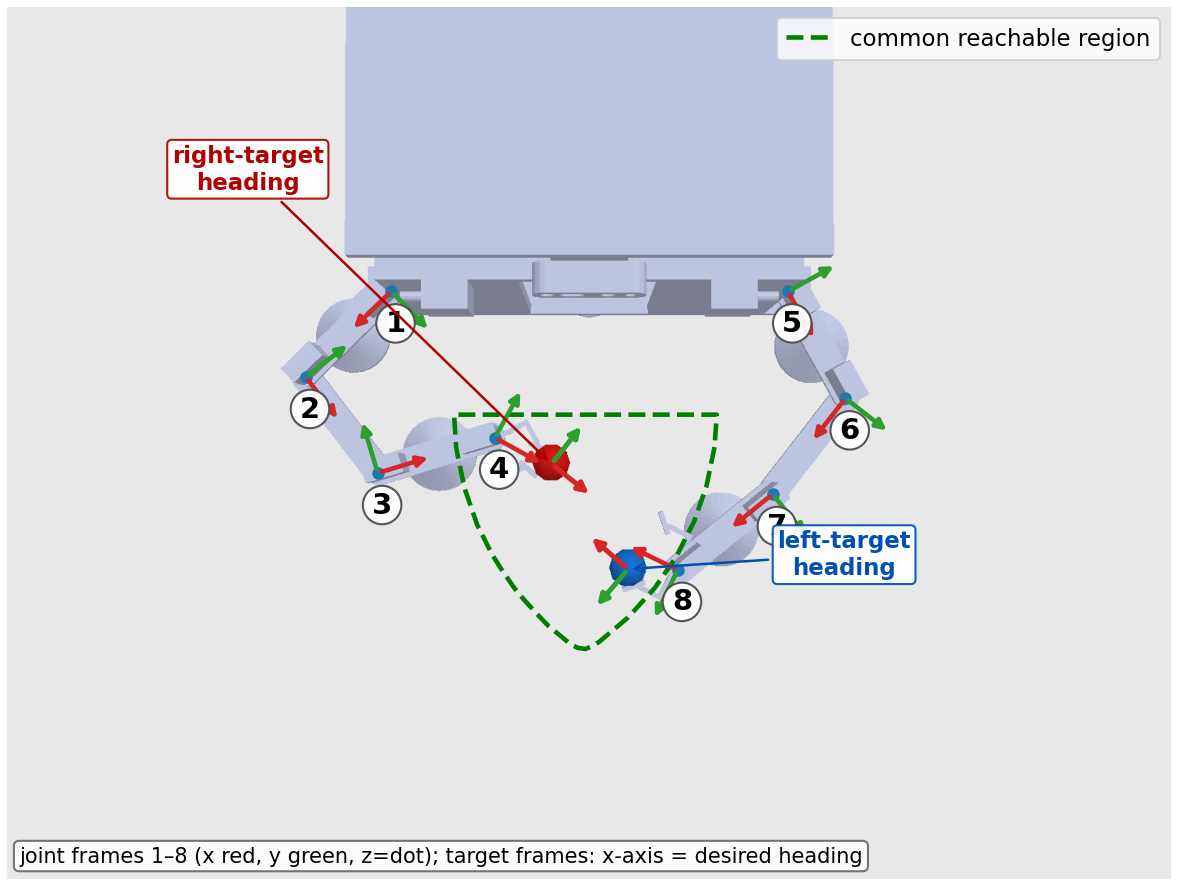}\\[2pt]
        (b)
	\end{minipage}
	\caption{Dual-arm space manipulator and simulation environment used in this study: (a) the simulation environment, showing the spacecraft base, the two four-joint arms with two-finger grippers, and the right (red) and left (blue) targets; (b) the kinematic schematic of the eight joint frames and the desired target headings within the common reachable region.}
	\label{fig:dual_arm_model}
\end{figure}

The dual-arm space manipulator considered in this study consists of two manipulators, each equipped with four revolute joints, three rigid links, and a two-finger gripper, as illustrated in Figure~\ref{fig:dual_arm_model}. This configuration supports repeated contact and reliable grasping while reducing the likelihood of inadvertently pushing the target away during cooperative manipulation. All motion is confined to a plane: each end-effector position is a two-dimensional coordinate and its orientation a single planar heading angle. The task is target reaching; a completed grasp is not required for success, and the grasp bonus introduced in Section~\ref{sec:problem_formulation} is an auxiliary shaping signal.


\begin{figure}[htb]
	\centering\includegraphics[width=3.5in]{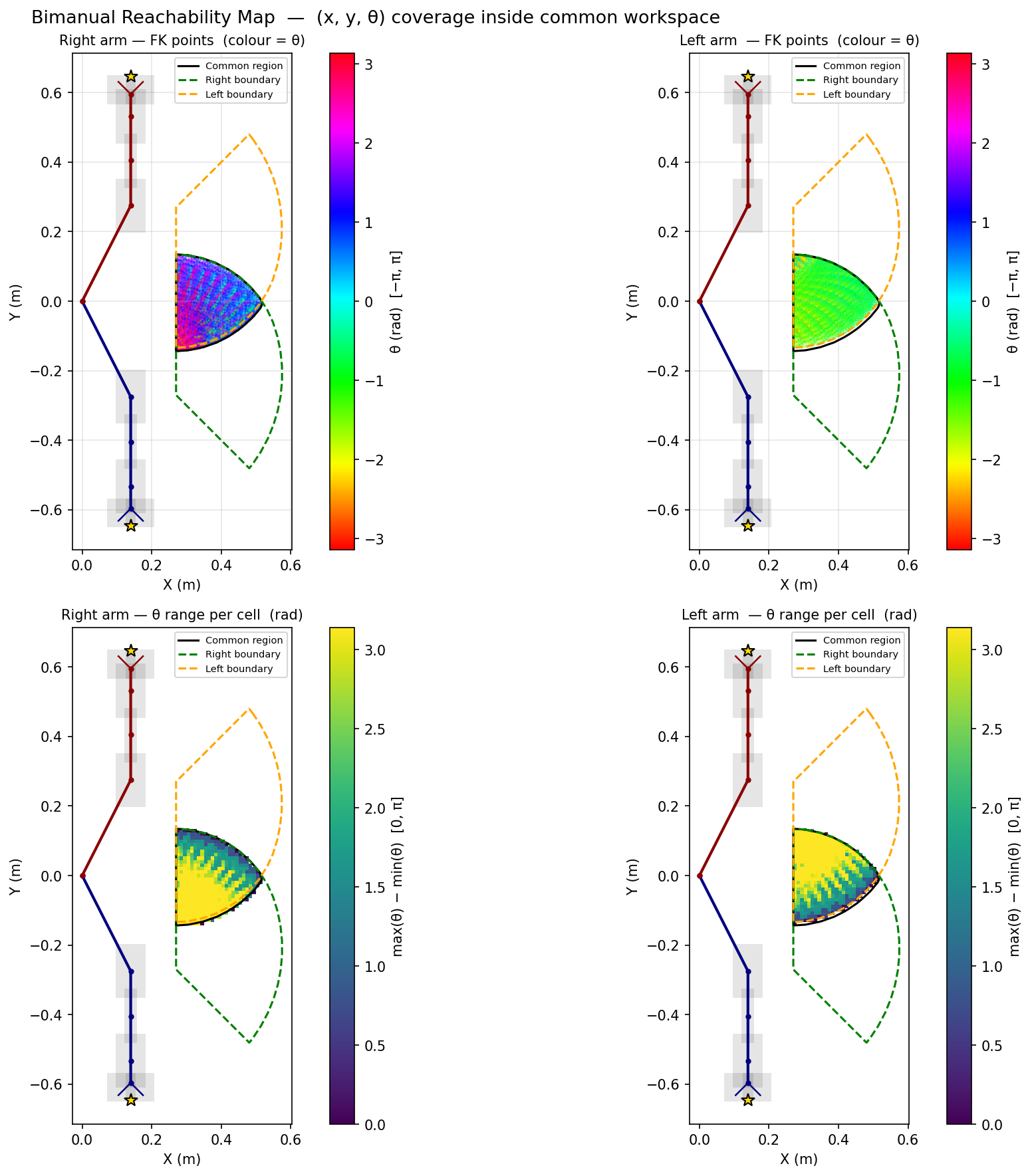}
	\caption{Bimanual reachability map of the dual-arm space manipulator.}
	\label{fig:reachability_map}
\end{figure}

Figure~\ref{fig:reachability_map} presents the bimanual reachability map of the dual-arm system, obtained by forward-kinematics sampling over the admissible joint space of each manipulator. The common reachable workspace, bounded by the solid black boundary, is the intersection of the individual workspaces of the right and left arms, shown by the green and orange dashed boundaries, respectively. Within this common workspace, the end-effector orientation coverage differs between the arms: the right arm attains a broad orientation range across most of the workspace, whereas the left arm exhibits more limited orientation freedom, particularly along the outer curved boundary where it approaches its maximum reach.

The reachability analysis indicates that the most kinematically constrained regions are concentrated near the outer boundary of the common workspace, especially for the left arm. Target configurations sampled near this boundary are therefore more likely to yield infeasible or marginally feasible orientation matching, which is consistent with the failure analysis reported in Section~\ref{sec:simulation_results}.
To mitigate this effect, target sampling is restricted to a conservative bimanual dexterous region in which both arms maintain an orientation range above a predefined threshold.
For simplicity, the spacecraft base is treated as fixed in this work. This assumption is reasonable when the base inertia is much larger than that of the manipulators, which reduces the dynamic coupling between arm and base motions. Under this assumption, the problem is formulated as a fixed-base dual-arm motion-planning task, while the extension to a free-floating base is left for future work.


%% file: 03_TechnicalPreliminaries.tex
\section{Technical Preliminaries}
\label{sec:technical_preliminaries}
This section formulates space-manipulator motion planning as a Markov decision process (MDP) for model-free deep reinforcement learning and reviews the two algorithms compared in this study. The MDP is defined as a tuple $\langle \mathcal{S}, \mathcal{A}, \mathcal{P}, \mathcal{R}, \gamma \rangle$, where $\mathcal{S}$ denotes the state space, $\mathcal{A}$ the action space, $\mathcal{P}$ the transition dynamics, $\mathcal{R}$ the reward function, and $\gamma$ the discount factor. Together, these elements define the agent--environment interaction and the balance between immediate and future rewards. The agent aims to learn a policy that maps observed states to actions so as to maximize the expected cumulative reward. In DRL, this policy is represented by a neural network and optimized through repeated interaction with the environment.

\subsection{Reinforcement Learning Algorithms}
On-policy methods such as PPO update only from data collected by the current policy, whereas off-policy methods such as SAC reuse stored past experience through a replay buffer. The former typically offer higher training stability, and the latter generally achieve better sample efficiency in high-dimensional continuous action spaces. To compare these two paradigms within an identical dual-arm task formulation, one representative algorithm from each category is investigated.

\subsection{Proximal Policy Optimization}

PPO~\cite{schulman2017proximal} is a policy-gradient method based on the actor--critic architecture that maximizes the expected cumulative reward while ensuring stable updates. The policy network $\pi_{\theta}$ outputs a probability distribution over actions, and the value network $V_{\phi}$ estimates the expected return. PPO optimizes the objective:
\begin{eqnarray}
J(\theta) = \mathbb{E}_{\tau \sim \pi_{\theta}(\tau)} \left[ \sum_{t} R(s_t, a_t) \right],
\end{eqnarray}
where $\tau$ denotes a trajectory of state--action pairs. The corresponding policy gradient is:
\begin{eqnarray}
\nabla_{\theta} J(\theta) = \mathbb{E}_{\tau \sim \pi_{\theta}(\tau)} \left[ \left( \sum_{t=1}^{T} \nabla_{\theta} \log \pi_{\theta}(a_t|s_t) \right) R(\tau) \right].
\end{eqnarray}

To reduce the high variance of gradient estimates, PPO adopts an actor--critic formulation using the advantage function $A^{\pi}(s, a) = Q^{\pi}(s, a) - V^{\pi}(s)$, which measures how much better action $a$ is relative to the average value of state $s$. Since $V^{\pi}(s)$ cannot be computed directly, the critic network approximates it from sampled trajectories.

The actor is updated by maximizing the clipped surrogate objective:
\begin{eqnarray}
L(\theta) = \hat{\mathbb{E}}_t \left[ \min \left( r_t(\theta) \hat{A}_t,\ \text{clip}\left(r_t(\theta), 1 - \epsilon, 1 + \epsilon \right) \hat{A}_t \right) \right],
\end{eqnarray}
where $r_t(\theta) = \pi_{\theta}(a_t \mid s_t) / \pi_{\theta_{\text{old}}}(a_t \mid s_t)$. Constraining this ratio within $[1-\epsilon, 1+\epsilon]$ prevents excessively large policy updates. The critic minimizes the TD value loss:
\begin{eqnarray}
L(\phi) = \hat{\mathbb{E}}_t \left[ \left( V_{\phi}(s_t) - R_t \right)^2 \right].
\end{eqnarray}

\subsection{Soft Actor-Critic}

SAC~\cite{haarnoja2018soft} augments the expected return with a policy-entropy term to encourage exploration in continuous action spaces. The entropy-regularized objective is
\begin{eqnarray}
J(\pi) = \mathbb{E}_{(s_t, a_t) \sim \rho_\pi} \left[ \sum_t \gamma^t \left( r(s_t, a_t) + \alpha \mathcal{H}\!\left(\pi(\cdot \mid s_t)\right) \right) \right],
\end{eqnarray}
where the entropy of a continuous policy is the expectation
\begin{eqnarray}
\mathcal{H}\!\left(\pi(\cdot \mid s_t)\right) = \mathbb{E}_{a_t \sim \pi}\left[ -\log \pi(a_t \mid s_t) \right],
\end{eqnarray}
and the temperature $\alpha$ trades off reward against entropy. This study uses the variant with twin soft $Q$-networks and automatic temperature tuning. Two critics $Q_{\theta_1}, Q_{\theta_2}$, with slowly updated target networks $Q_{\bar{\theta}_1}, Q_{\bar{\theta}_2}$, are trained to minimize the soft Bellman error
\begin{eqnarray}
J_Q(\theta_i) = \mathbb{E}_{(s_t, a_t) \sim \mathcal{D}} \left[ \tfrac{1}{2}\left( Q_{\theta_i}(s_t, a_t) - y_t \right)^2 \right], \quad i \in \{1,2\},
\end{eqnarray}
with the target
\begin{eqnarray}
y_t = r(s_t, a_t) + \gamma\, \mathbb{E}_{a_{t+1} \sim \pi_\phi} \left[ \min_{j=1,2} Q_{\bar{\theta}_j}(s_{t+1}, a_{t+1}) - \alpha \log \pi_\phi(a_{t+1} \mid s_{t+1}) \right],
\end{eqnarray}
where $\mathcal{D}$ is the replay buffer and the minimum over the two critics mitigates overestimation bias. The policy $\pi_\phi$ is updated to maximize the entropy-regularized soft value:
\begin{eqnarray}
J_\pi(\phi) = \mathbb{E}_{s_t \sim \mathcal{D},\, a_t \sim \pi_\phi} \left[ \alpha \log \pi_\phi(a_t \mid s_t) - \min_{j=1,2} Q_{\theta_j}(s_t, a_t) \right].
\end{eqnarray}
The temperature is adjusted automatically to keep the policy entropy near a target value $\bar{\mathcal{H}}$ by minimizing
\begin{eqnarray}
J(\alpha) = \mathbb{E}_{a_t \sim \pi_\phi} \left[ -\alpha \left( \log \pi_\phi(a_t \mid s_t) + \bar{\mathcal{H}} \right) \right].
\end{eqnarray}
Reusing stored transitions from the replay buffer $\mathcal{D}$ is the source of SAC's sample efficiency and, as discussed in Section~\ref{sec:simulation_results}, of its ability to amplify rare successful transitions. The parameter subscripts $\theta$ and $\phi$ are local to each algorithm subsection.

%% file: 04_ProblemFormulation.tex
\section{Problem Formulation}
\label{sec:problem_formulation}

\subsection{Markov Decision Process Formulation}
Building on the MDP defined in Section~\ref{sec:technical_preliminaries}, the dual-arm space-manipulator motion-planning problem is instantiated through the agent--environment interaction, comprising the state representation, action space, transition dynamics, and reward function. This formulation provides the basis for learning coordinated motion policies that enable both manipulators to reach their assigned targets while satisfying task-level and safety constraints.

\subsection{State and Action Spaces}
The centralized state representation includes the joint angles of each arm, the end-effector positions and velocities, the relative position between each end-effector and its target, and the heading error of each arm. These components collectively define a state vector $s \in \mathbb{R}^{24}$, organized into six groups:
\begin{eqnarray}
s = \left[ \theta_r,\ \theta_l,\ p_r^{ee},\ p_l^{ee},\ \Delta p_r,\ \Delta p_l,\ v_r^{ee},\ v_l^{ee},\ \psi_r,\ \psi_l,\ \Delta\psi_r,\ \Delta\psi_l \right],
\end{eqnarray}
where $\theta_r, \theta_l \in \mathbb{R}^4$ are the absolute joint angles of the right and left arms; $p_r^{ee}, p_l^{ee} \in \mathbb{R}^2$ are the right and left end-effector XY positions in the world frame; $\Delta p_r = p_r^{target} - p_r^{ee}$ and $\Delta p_l = p_l^{target} - p_l^{ee} \in \mathbb{R}^2$ are the target-relative positions between each end-effector and its target; $v_r^{ee}, v_l^{ee} \in \mathbb{R}^2$ are the right and left end-effector velocities; $\psi_r, \psi_l \in \mathbb{R}$ are the right and left end-effector orientations; and $\Delta\psi_r, \Delta\psi_l \in \mathbb{R}$ are the right and left heading errors, computed as the normalized angular difference between the target orientation and the current end-effector heading.

The action space has dimension $a \in \mathbb{R}^{8}$, where each action corresponds to an incremental change in a joint angle, representing a relative adjustment from the current configuration rather than an absolute joint position. The action vector is defined as:
\begin{eqnarray}
a = \left[ \Delta\theta_{r,0:3},\ \Delta\theta_{l,0:3} \right],
\end{eqnarray}
where $\Delta\theta_{r,0:3} \in \mathbb{R}^4$ and $\Delta\theta_{l,0:3} \in \mathbb{R}^4$ denote the incremental joint-angle adjustments (rad) applied to the right and left arms, respectively, at each control step.

\subsection{Transition Dynamics}
The normalized policy output $a \in [-1, 1]^8$ is scaled by $0.05$ and added to the current joint angles, so each control step applies an incremental change of at most $0.05$~rad per joint; the updated angles are clamped to the joint limits of each manipulator. Each control action is held for four simulation substeps before the next observation is taken. The dynamics are deterministic: no process or observation noise is injected in the experiments reported here. The success and collision conditions, together with the episode horizon $T_{\max} = 1000$ control steps, are specified in Section~\ref{sec:problem_formulation} (Episode Termination and Success Criteria).

\subsection{Reward Function Design}
To train the centralized policy for cooperative dual-arm manipulation, a multi-objective reward function is designed to promote accurate target reaching, orientation alignment, smooth motion, inter-arm collision avoidance, and timely task completion. The total reward at each time step is defined as:
\begin{eqnarray}
r_t = r_{\text{task}} + r_{\text{smooth}} + r_{\text{prox}} + r_{\text{collision}} + r_{\text{grasp}} + r_{\text{async}} + r_{\text{time}} + r_{\text{success}},
\end{eqnarray}
where each term corresponds to a distinct behavioral objective, described below.

\emph{Two-phase task reward.} The primary task reward $r_{\text{task}}$ is decomposed into two sequential phases for each arm, mirroring the two-stage nature of the manipulation task: first reaching the target position, then aligning the end-effector orientation. In Phase~1 (position not yet reached), the reward is dominated by a position term that drives the end-effector toward the target, supplemented by a small proximity-weighted orientation hint that biases the approach direction. In Phase~2 (the end-effector has entered the success region), the position term is disabled and a stronger orientation term takes over, providing a clear signal for wrist rotation. For arm $i \in \{r, l\}$, this is formulated as:
\begin{eqnarray}
r_{\text{task}}^{i} =
\begin{cases}
2.0 \left(1 - \tanh(10\, d_i)\right) + 0.5\, e^{-d_i / 0.25} \left(1 - \tanh(\Delta\psi_i)\right), & \text{if position not reached}, \\[4pt]
2.0 \left(1 - \tanh(2\, \Delta\psi_i)\right), & \text{if position reached},
\end{cases}
\end{eqnarray}
where $d_i$ denotes the Euclidean distance between end-effector $i$ and its target, and $\Delta\psi_i$ denotes the corresponding heading error. The total task reward is $r_{\text{task}} = r_{\text{task}}^{r} + r_{\text{task}}^{l}$, with each arm's contribution disabled once that arm's sub-task is complete. This staged formulation ensures the agent first prioritizes spatial alignment before committing to orientation refinement, avoiding premature or unstable wrist rotations far from the target.

\emph{Motion smoothness penalty.} To penalize erratic or high-frequency motion, a smoothness penalty $r_{\text{smooth}}$ is applied based on end-effector velocity, joint velocity, and joint acceleration. The end-effector velocity penalty discourages excessive linear speed of the end-effector:
\begin{eqnarray}
r_{\text{vel}}^{i} = -0.05 \, \tanh\!\left(3.0 \max(0,\, \|v_i^{ee}\| - 0.5)\right),
\end{eqnarray}
while joint-level jitter is penalized using the mean absolute joint velocity:
\begin{eqnarray}
r_{\text{jitter}}^{i} = -0.02 \, \tanh\!\left(2.0 \max(0,\, \bar{\omega}_i - 1.0)\right),
\end{eqnarray}
and joint acceleration, approximated via the change in joint velocity between consecutive steps, is penalized to further promote smooth trajectories:
\begin{eqnarray}
r_{\text{acc}}^{i} = -0.01 \, \min\!\left(1.0,\, \overline{|\Delta\omega_i|}\right).
\end{eqnarray}
The total smoothness penalty is the sum of these three terms across both arms,
\begin{equation}
    r_{\text{smooth}} = \sum_{i \in \{r,l\}} \left( r_{\text{vel}}^{i} + r_{\text{jitter}}^{i} + r_{\text{acc}}^{i} \right).
\end{equation}

\emph{Non-end-effector proximity penalty.} To penalize the robot's body links (excluding the end-effector and fingertips) from approaching the target, a proximity penalty $r_{\text{prox}}$ is applied based on the minimum distance between any non-end-effector link and the nearest target:
\begin{eqnarray}
r_{\text{prox}} = -0.5 \left(1 - \tanh\!\left(15.0 \, d_{\min}^{\text{link}}\right)\right),
\end{eqnarray}
where $d_{\min}^{\text{link}}$ denotes the minimum XY distance between any forearm link and the active targets. This term encourages the manipulator to approach using its end-effector rather than through incidental contact from the arm body.

\emph{Self-collision penalty.} A smooth, bounded collision penalty $r_{\text{collision}}$ is applied when the minimum inter-link distance falls below a safety threshold $d_{\text{safe}} = 0.02$~m:
\begin{eqnarray}
r_{\text{collision}} =
\begin{cases}
-0.5 \left(1 - \tanh\!\left(k_c \, \dfrac{d_{\min}}{d_{\text{safe}}}\right)\right), & \text{if } d_{\min} < d_{\text{safe}}, \\[6pt]
0, & \text{otherwise},
\end{cases}
\end{eqnarray}
where $d_{\min}$ is the minimum distance between robot links and $k_c = 30.0$ controls the steepness of the penalty. The $\tanh$ formulation provides a smooth, differentiable signal that increases sharply as the arms approach self-collision, while avoiding the instability of a hard penalty threshold.

\emph{Grasp and coordination terms.} A sparse grasp bonus $r_{\text{grasp}}$ of $+1.0$ is awarded to each arm independently upon detecting valid finger contact with its assigned target, providing an immediate positive signal for successful grasping. To promote coordinated bimanual completion, an asymmetric progress penalty $r_{\text{async}} = -0.1$ is applied at every step in which one arm has completed its sub-task while the other has not, discouraging the policy from neglecting one arm in favor of the other.

\emph{Time and success rewards.} A small constant efficiency penalty of $-0.005$ per step, together with a penalty that scales linearly with episode progress, $-0.01 \, (t / T_{\max})$, is applied to encourage faster task completion as the episode advances. Finally, a single large sparse bonus $r_{\text{success}} = +300.0$ is awarded upon successful task termination. Unlike earlier formulations that granted smaller per-step bonuses for each completed sub-task, this reward is given only once at episode completion; per-step completion bonuses were found to induce a reward-loitering behavior in which the policy oscillated near the success boundary to repeatedly collect partial rewards without finishing the episode.

\subsection{Episode Termination and Success Criteria}
An episode terminates upon successful task completion or after a maximum of $T_{\max} = 1000$ control steps. An arm is considered successful when the target lies inside the end-effector grip circle (the circular position-tolerance region, which tightens from $2.5$~cm on approach to $1.0$~cm for final completion) and its heading error satisfies $\Delta\psi \le 15^\circ$. The episode succeeds when both arms are simultaneously latched, meaning that both maintain their position and heading tolerances, before $T_{\max}$. These success conditions coincide with the Phase-2 trigger of the two-phase task reward defined above, ensuring consistency between the reward and the reported success metric. A self-collision is registered whenever the minimum inter-link clearance falls below the safety threshold $d_{\text{safe}} = 0.02$~m used by the collision penalty.

%% file: 05_SimulationResults.tex
\section{Simulation Results}
\label{sec:simulation_results}
To assess the effectiveness of the proposed framework and to compare on-policy and off-policy learning for the target-reaching task with inter-arm collision avoidance, PPO and SAC are evaluated within the same task formulation. The objective is not only to verify that the framework can learn the cooperative motion-planning task, but also to examine its generalization to previously unseen target configurations. Unless stated otherwise, all comparative statements are scoped to the considered fixed-base dual-arm simulation scenarios.

\subsection{Simulation Setup}
The environment is implemented in PyBullet and the policies are trained with Stable-Baselines3. Each episode is capped at $T_{\max} = 1000$ control steps. During training, targets are resampled every 10 episodes from the interior of the common reachable region (with a 3~cm erosion of the boundary) with inter-target separation $\sim\mathcal{U}(0.07, 0.20)$~m; during evaluation, targets are resampled every episode from a fixed, shared seed sequence so that episode-level comparisons across models are paired. Success and collision are defined as in Section~\ref{sec:problem_formulation}: both arms must satisfy the position and heading tolerances before $T_{\max}$, and an episode is flagged as a collision when the minimum inter-link clearance falls below $d_{\text{safe}} = 0.02$~m. All ten runs (five seeds per algorithm) share an identical MDP configuration (verified by a common configuration hash), so that only the algorithm-side hyperparameters differ. The simulation environment is shown in Figure~\ref{fig:dual_arm_model}(a).

\subsection{Training Configuration}
The final hyperparameters for PPO and SAC are summarized in Table~\ref{tab:training_hyperparameters}. Both algorithms are trained for an identical budget of 4M environment steps over five random seeds; the fair-comparison currency is environment steps rather than wall-clock time or episodes. Following best practices for reproducible deep reinforcement learning evaluation~\cite{henderson2018deep,agarwal2021deep}, per-seed point estimates are aggregated across seeds and reported with a bootstrap 95\% confidence interval (10{,}000 resamples), using the across-seed mean for the success and collision rates and the across-seed median for the continuous error metrics; headline PPO-vs-SAC gaps are assessed with the Mann--Whitney $U$ test and Cliff's $\delta$ effect size. Because $n = 5$ yields low statistical power, differences are reported as distinguishable/indistinguishable at $n = 5$ and supported by non-overlapping confidence intervals rather than by a $p$-value alone.

Both algorithms optimize the identical MDP (verified by the shared configuration hash), so the differences in Table~\ref{tab:training_hyperparameters} are algorithm-side; two are disclosed for completeness. First, PPO uses observation and reward normalization (with clipping at $\pm 10$), a standard preprocessing step for on-policy PPO, whereas SAC uses raw observations; this normalization preserves the ratio between the terminal success bonus and the dense shaping reward, and the normalized success bonus remains well within the clipping range. Second, the network sizes, batch sizes, and number of parallel environments follow common per-algorithm defaults. The hyperparameters were obtained by manual tuning of each algorithm rather than a systematic equal-budget search, which we note as a limitation of the comparison.

\begin{table}[htbp]
\centering
\caption{Final training hyperparameters for PPO and SAC. GAE: generalized advantage estimation.}
\label{tab:training_hyperparameters}
\begin{tabular}{lcc}
\hline
Parameter & PPO & SAC \\
\hline
Learning rate & $1\times10^{-4}$ (decayed) & $3\times10^{-4}$ \\
Discount factor $\gamma$ & 0.99 & 0.97 \\
Network architecture & $[512, 512]$ & $[512, 512, 256]$ \\
Batch size & 512 & 256 \\
Rollout length $n_{\text{steps}}$ & 2048 & N/A \\
Clip range & 0.15 (decayed) & N/A \\
GAE $\lambda$ & 0.95 & N/A \\
Entropy coefficient & 0.005 & auto ($\bar{\mathcal{H}} = -8$) \\
Replay buffer size & N/A & $10^{6}$ \\
Training steps & 4M & 4M \\
\hline
\end{tabular}
\end{table}

\subsection{Training Convergence}
\begin{figure}[htb]
	\centering
	\includegraphics[width=3.3in]{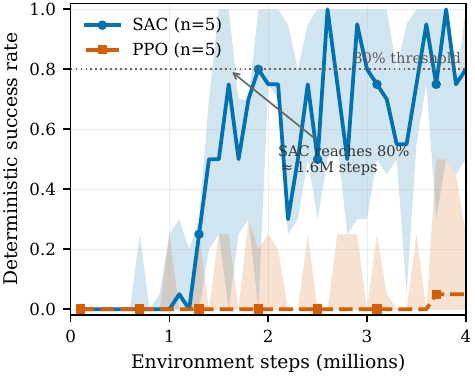}
	\caption{Training convergence of PPO and SAC: deterministic success rate versus environment steps (median with bootstrap 95\% CI over five seeds).}
	\label{fig:training_curves}
\end{figure}

The learning curves in Figure~\ref{fig:training_curves} report the deterministic success rate, the policy evaluated with exploration disabled, using the mean action, versus environment steps (median $\pm$ bootstrap 95\% CI over five seeds). SAC reaches the 80\% success threshold in all five seeds, at a median of 1.6M steps [1.5M, 3.0M], whereas PPO does not reach that threshold in any seed within the 4M-step budget (a right-censored outcome, not a finite value). The normalized area under the success-versus-steps curve is 0.497 [0.426, 0.510] for SAC and 0.032 [0.019, 0.052] for PPO.

\begin{figure}[htb]
	\centering
	\includegraphics[width=3.3in]{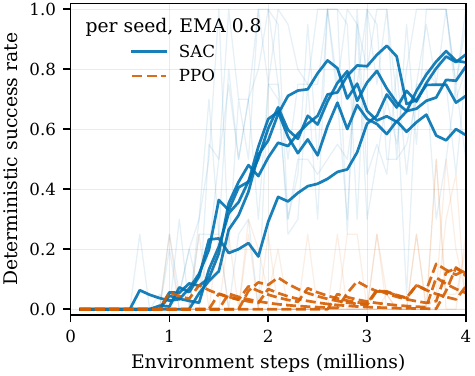}
	\caption{Per-seed deterministic success rate versus environment steps for all five seeds of PPO and SAC (faint lines: raw checkpoints; bold lines: exponential moving average, weight 0.8).}
	\label{fig:training_curves_seeds}
\end{figure}

Figure~\ref{fig:training_curves_seeds} shows the individual per-seed success curves, smoothed with an exponential moving average (weight 0.8), in the style of a TensorBoard visualization. All five SAC seeds rise to a substantial success level, whereas all five PPO seeds remain near zero, confirming that the separation is consistent across seeds rather than driven by an outlier.

\subsection{Task Performance}
\begin{figure}[htb]
	\centering
	\includegraphics[width=\linewidth]{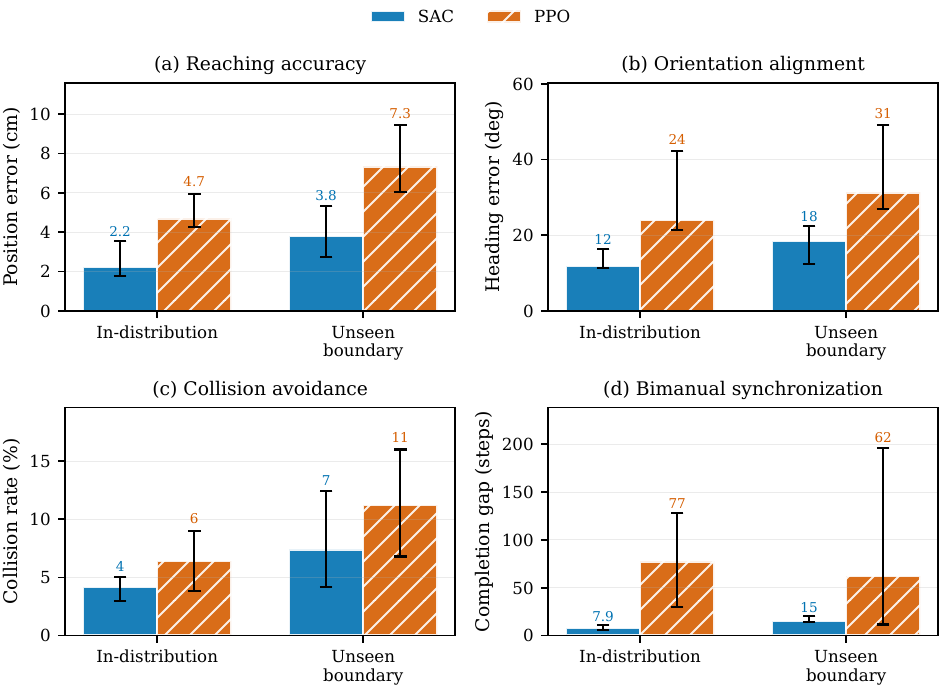}
	\caption{Per-objective task performance for PPO and SAC, in-distribution and on unseen boundary targets (bootstrap 95\% CI over five seeds; the collision rate is an across-seed mean, the other metrics across-seed medians): (a) reaching accuracy, (b) orientation alignment, (c) collision avoidance, and (d) bimanual synchronization.}
	\label{fig:task_metrics}
\end{figure}

Table~\ref{tab:task_performance} reports the in-distribution task-performance metrics for the trained policies (100 deterministic episodes per seed). SAC attains a mean success rate of 78.6\% [72.6, 84.2] versus 7.4\% [3.2, 11.4] for PPO ($p = 0.0079$, the minimum attainable at $n = 5$; Cliff's $\delta = +1.0$, i.e., every SAC seed exceeds every PPO seed). SAC also achieves lower final position error (2.24~cm vs.\ 4.68~cm), lower heading error ($11.9^\circ$ vs.\ $24.0^\circ$), and a lower collision rate (4.2\% vs.\ 6.4\%). The collision rate is measured per episode; because PPO's episodes are far longer (569 vs.\ 123 steps, Table~\ref{tab:task_performance}), this compares safety per episode rather than per unit of motion. The per-objective results are visualized in Figure~\ref{fig:task_metrics}.

\begin{table}[htbp]
\centering
\caption{In-distribution task performance over five seeds (100 episodes per seed) [bootstrap 95\% CI]; success and collision rates are across-seed means, the remaining metrics across-seed medians. Bold marks the better value.}
\label{tab:task_performance}
\begin{tabular}{lcc}
\hline
Metric & SAC & PPO \\
\hline
Success rate (\%) & \textbf{78.6 [72.6, 84.2]} & 7.4 [3.2, 11.4] \\
Final position error (cm) & \textbf{2.24 [1.80, 3.54]} & 4.68 [4.25, 5.95] \\
Final heading error (deg) & \textbf{11.9 [11.3, 16.3]} & 24.0 [21.4, 42.2] \\
Collision rate (\%) & \textbf{4.2 [3.0, 5.0]} & 6.4 [3.8, 9.0] \\
Episode length (steps) & \textbf{123 [101, 135]} & 569 [500, 591] \\
Inter-arm completion gap (steps) & \textbf{7.9 [5.8, 11.2]} & 77.3 [30.0, 127.7] \\
Mean $|$joint acceleration$|$ (rad/s/step) & 0.080 [0.076, 0.081] & 0.046 [0.031, 0.063] \\
Mean end-effector speed (m/s) & 0.514 [0.495, 0.523] & 0.287 [0.249, 0.356] \\
\hline
\end{tabular}
\end{table}

The synchronization gap is large: SAC completes both arms within a median of about 8 steps of each other, versus about 77 steps for PPO, supporting the synchronized-bimanual-completion objective. This gap is computed only over episodes in which both arms latch, so PPO's value reflects its rare near-successful episodes. PPO's long episodes (569 vs.\ 123 steps) and higher raw return (1424 vs.\ 347) reflect a reward-loitering local optimum, in which the policy hovers a few centimeters from the targets to harvest dense shaping reward rather than committing to the precision latch. Its lower mean joint acceleration and end-effector speed therefore reflect reduced motion from hovering rather than superior smoothness, and should be read together with episode length. Per-seed success rates (SAC 84/85/69/72/83\%, PPO 10/1/3/14/9\%) confirm that the separation is consistent across seeds.

\subsection{Generalization to Unseen Target Configurations}
\begin{figure}[htb]
	\centering
	\includegraphics[width=3.3in]{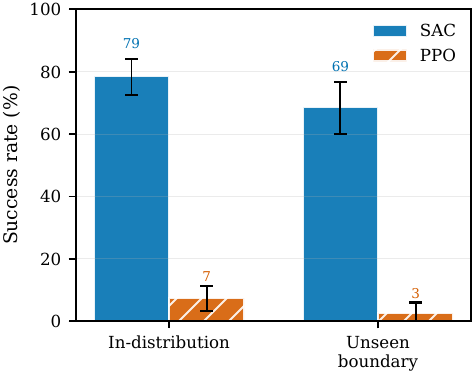}
	\caption{Success rate of PPO and SAC on in-distribution and unseen boundary targets (across-seed mean with bootstrap 95\% CI over five seeds).}
	\label{fig:generalization}
\end{figure}

Generalization is evaluated on a held-out distribution of targets drawn from the outer 3~cm band of the reachable region, which the interior-eroded training sampler never visits and which is therefore spatially disjoint from the training distribution. As shown in Table~\ref{tab:generalization} and Figure~\ref{fig:generalization}, both algorithms degrade out of distribution, but to very different degrees: SAC retains about 87\% of its in-distribution success (68.6\% vs.\ 78.6\%), whereas PPO's already-low competence largely collapses (2.6\% vs.\ 7.4\%). Position and collision metrics worsen for both algorithms near the workspace boundary, consistent with the reduced orientation feasibility identified in the reachability analysis of Section~\ref{sec:proposed_system}. The success gap remains distinguishable at $n = 5$ ($\delta = +1.0$; the exact test gives $p = 0.0079$, as in the in-distribution case, since the two groups are completely separated).

\begin{table}[htbp]
\centering
\caption{Generalization to unseen boundary targets over five seeds [bootstrap 95\% CI]; success and collision rates are across-seed means, the remaining metrics across-seed medians. Bold marks the better value.}
\label{tab:generalization}
\begin{tabular}{lcc}
\hline
Metric & SAC & PPO \\
\hline
Success rate (\%) & \textbf{68.6 [60.0, 76.8]} & 2.6 [0.2, 6.0] \\
Final position error (cm) & \textbf{3.83 [2.72, 5.33]} & 7.32 [6.06, 9.43] \\
Final heading error (deg) & \textbf{18.5 [12.5, 22.4]} & 31.2 [27.1, 49.3] \\
Collision rate (\%) & \textbf{7.4 [4.2, 12.4]} & 11.2 [6.8, 16.0] \\
\hline
\end{tabular}
\end{table}

\subsection{PPO Failure-Mode Controls}
To identify why PPO fails, we examine the in-distribution episodes and run three control experiments. A success funnel over the 500 in-distribution episodes (five seeds pooled) shows that PPO does not fail at reaching: about half of its episodes bring both end-effectors within 5~cm of the targets. Instead, it attrits at the precision latch: only 14.8\% of PPO episodes ever latch both arms, versus 79.8\% for SAC, and two-thirds of one-arm latches never complete the second arm, consistent with the large inter-arm completion gap. The following control experiments (single seed) isolate candidate causes.

\emph{Discount.} Because PPO uses a higher discount factor ($\gamma = 0.99$) than SAC ($\gamma = 0.97$), the discounted value of prolonged loitering is larger for PPO, raising the concern that its behavior is merely a rational response to its own discount. Retraining PPO with $\gamma = 0.97$ and everything else identical leaves success at 4\% (versus 10\% for the same seed at $\gamma = 0.99$), so equalizing the discount does not recover PPO.

\emph{Training budget.} Extending PPO training well beyond the 4M-step budget does not change the outcome: a single-seed run plateaus at 15--22\% success by 16M steps, and the best configuration reaches only 1\% success at 50M steps ($12.5\times$ the budget), with deterministic checkpoint success at zero throughout the second half of training. PPO's result is a converged plateau, not undertraining.

\emph{Reward shaping.} Replacing the dense shaped reward with a potential-based reward that removes the per-step loiter income lowers success for both algorithms but does not close the gap: SAC drops from 84\% to 54\% (still solving the task) while PPO drops from 10\% to 0\%. The separation is therefore robust to the reward design.

Together, these controls indicate that PPO's failure is not explained by the discount, the training budget, or the specific reward shaping. Rather, the shaped reward admits a high-return loitering optimum, and the successful precision latches occur at a near-zero rate in the rollout data. PPO's on-policy updates, which reuse each batch for only a few epochs before discarding it, cannot amplify these rare events, whereas SAC's replay buffer reuses each rare success many times and propagates the terminal success bonus through its $Q$-function. The single-seed controls are supplementary evidence; the headline comparison remains the five-seed results above.

\subsection{Discussion}
Within the considered fixed-base dual-arm simulation scenarios, SAC converges faster (reaching 80\% success in all five seeds while PPO never reaches it within 4M steps), attains substantially higher task success, produces more accurate and better-synchronized motions, and generalizes far better to unseen boundary targets than PPO. These tendencies are consistent with the reported stability advantage of maximum-entropy off-policy methods on manipulation benchmarks~\cite{xiang2023rmbench} and with PPO's known sensitivity to hyperparameter tuning~\cite{shianifar2024optimizing}; however, the differences reported here are attributed to the present multi-seed measurements rather than to any universal ordering of the two algorithms.

PPO's failure mode is specific and should not be read as general PPO inferiority: as the control experiments above show, PPO converges to a reward-loitering local optimum that persists across discount, budget, and reward-shaping variations. Because episode return is not monotone in task performance under this reward, success rate is used as the primary metric throughout.

\emph{Limitations.} The study is confined to a fixed-base formulation evaluated only in simulation, with $n = 5$ seeds per algorithm. Robustness to observation noise, listed among the target objectives, was not evaluated in this study and is left for future work, as are the extension to a free-floating base and sim-to-real transfer.

%% file: 06_Conclusion.tex
\section{Conclusion}
\label{sec:conclusion}
This paper presented a coordinated reinforcement learning framework for cooperative motion planning of a dual-arm space manipulator in a target-reaching task with inter-arm collision-avoidance constraints. The proposed formulation combined a centralized state representation with a multi-objective reward function to promote target-reaching accuracy, orientation alignment, smooth motion, synchronized task completion, and inter-arm collision avoidance. PPO and SAC were compared as representative on-policy and off-policy algorithms within the same task formulation, and the learned policies were further examined for generalization to previously unseen target configurations.

Across five random seeds, SAC substantially outperformed PPO within the considered fixed-base simulation scenarios, reaching an 80\% success threshold in every seed (median 1.6M environment steps) while PPO did not reach it within a 4M-step budget, and achieving a higher mean success rate (78.6\% versus 7.4\%) together with more accurate and better-synchronized motions. On unseen boundary targets, SAC retained most of its performance whereas PPO's largely collapsed. The comparison also revealed that, under the two-phase shaped reward, PPO tends to converge to a reward-loitering local optimum. Control experiments showed that this behavior persisted under a matched discount factor, a $12.5\times$ larger training budget, and an alternative reward design, pointing to a limitation of on-policy sample reuse rather than a general deficiency of PPO.

Future work will evaluate robustness to observation noise, which was not assessed in this study, and will extend the current fixed-base formulation to free-floating space-manipulator dynamics, incorporate more realistic contact and grasping interactions, and investigate sim-to-real transfer and hardware validation.

%% file: 07_Acknowledgment.tex
\section{Acknowledgment}
This work was supported by the Discovery Grant (RGPIN-2024-06290) and the Collaborative Research and Training Experience Program Grant—SMART-ART (555425-2021) from the Natural Sciences and Engineering Research Council of Canada.
